\documentclass[10pt,twocolumn,letterpaper]{article}

\usepackage{iccv}
\usepackage{times}
\usepackage{epsfig}
\usepackage{graphicx}
\usepackage{amsmath}
\usepackage{amssymb}

\usepackage{multirow}
\usepackage{xcolor} 
\usepackage{subfiles}  
\usepackage{subfigure} 
\usepackage{array} 
\usepackage{booktabs} 

\usepackage{xcolor} 
\newcommand{\sgg}[1]{{\color{black}#1}}

\newcommand{\jy}[1]{{\color{black}#1}}
\newcommand{\jl}[1]{{\color{black}#1}}

\usepackage[breaklinks=true,bookmarks=false]{hyperref}

\iccvfinalcopy 

\def\iccvPaperID{****} 

\ificcvfinal\fi

\begin{document}

\title{GridCLIP: One-Stage Object Detection 

by Grid-Level CLIP Representation Learning}

\author{Jiayi Lin\\
Queen Mary University of London\\
{\tt\small jiayi.lin@qmul.ac.uk}
\and
Shaogang Gong\\
Queen Mary University of London\\
{\tt\small s.gong@qmul.ac.uk}
}

\maketitle
\ificcvfinal\thispagestyle{empty}\fi

\begin{abstract}
A vision-language foundation model pretrained on very large-scale image-text paired data 
has the potential to provide generalizable knowledge representation for
downstream visual recognition and detection tasks,  especially on
  supplementing the undersampled categories in downstream  model training.
Recent studies utilizing
    CLIP for  object detection have shown 
    that a two-stage detector design typically 
    outperforms a one-stage detector, 
    while requiring more expensive
    training resources and longer inference time.
 In this work, we propose a one-stage detector \jy{GridCLIP}
    \sgg{that narrows its} \jy{performance} gap \sgg{to those of} two-stage detectors,
    with approximately 43$\times$ and 5$\times$ faster than its two-stage counterpart (ViLD) in the training and test process respectively.
 \jy{GridCLIP learns grid-level representations 
    to adapt to the intrinsic \sgg{principle of one-stage detection learning}
    by expanding}
    the conventional CLIP image-text holistic mapping 
    to a more fine-grained, grid-text alignment.
    \sgg{This differs}
    \jy{from the region-text mapping 
    in two-stage detectors \sgg{that apply} CLIP directly by
    treating regions as images.}
Specifically, GridCLIP \jy{performs Grid-level Alignment to adapt} the CLIP image-level representations 
    to grid-level representations 
    \jy{by aligning} to CLIP category representations to learn the annotated (especially frequent) categories.
To learn generalizable visual representations of broader 
    categories, especially undersampled ones,
    we perform Image-level Alignment during training to
    \sgg{propagate broad pre-learned categories in the CLIP image encoder from
        the image-level to the grid-level representations}.
Experiments show that the learned CLIP-based grid-level representations boost the performance of undersampled (infrequent and novel) categories, 
reaching \jy{comparable} detection performance on the LVIS benchmark.
\end{abstract}


\section{Introduction}
\begin{figure}[ht]
   \centering
   \includegraphics[width=7.2cm]{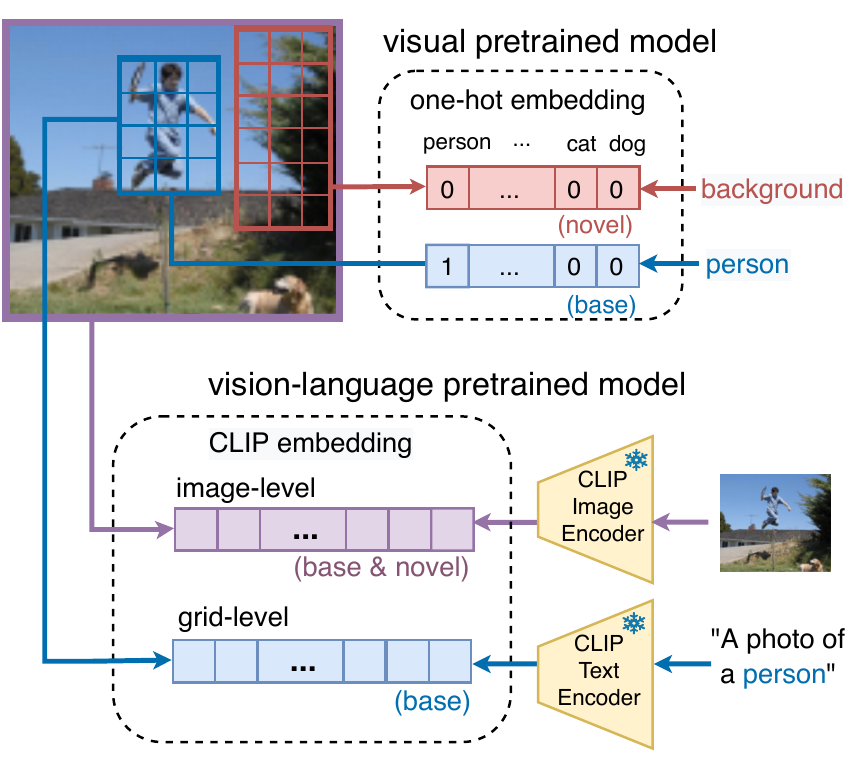}
   \caption{ Comparison of approaches to applying visual pretrained and vision-language pretrained models (GridCLIP as an example) 
   to one-stage detectors.
   The visual pretrained model is commonly used for extracting  grid-level image representations (also called feature maps) which are aligned to the manual one-hot embeddings, so only base categories can be learned (top). While in a vision-language pretrained model, images are encoded into high-dimension embeddings, which can be aligned to the text embeddings of base categories as well as the whole image embedding of both base and novel categories (bottom). 
   \label{intro}}
\end{figure}
Simultaneous multi-category object detection aims to 
both recognize (classify) and detect (locate) all instances of  given
  categories in an image. 
  A  significant challenge in training a good detector
is the  cost of labeling a large-scale dataset on a broad
  range of object categories with balanced data distributions. Existing
  detection datasets are often
  imbalanced with a long-tail distribution across
  categories~\cite{saichev2009theory} where some object categories have only
  a few or zero training sample(s).
To deal with these undersampled categories, 
    few-shot and zero-shot learning  have been explored, but
      they are inherently weaker models when compared to a fully supervised
      learning based model. 
      
\jy{Elsewhere, Self-Supervised Learning (SSL) has received increasing research interest for} exploring widely available unlabelled data.
 Self-supervised pretraining followed by supervised
  fine-tuning for constructing a detector has been proposed recently~\cite{yang2021instance, wei2021aligning}.
 An example is Open Vocabulary Object Detection (OVOD)~\cite{zareian2021open}, 
    which pretrains a model on image-caption pairs containing a
    substantial amount of  broad categories, 
     followed by fine-tuning the model on a detection
     specific dataset of only a few base categories.
 For the pretraining stage that supplements knowledge for undersampled categories,  Vision-Language pretrained Models  (VLMs) are widely adopted.
As one of the most widely adopted VLMs, CLIP~\cite{radford2021learning} is pretrained 
    on a dataset of 400 million image-text pairs  with a vocabulary of over 49,000 words, providing generalizable visual embeddings 
    of broad categories
    that helps supplement the undersampled categories in a detection dataset.



Recent approaches have been exploring the CLIP-based representation for object detection,  mostly for Open-Vocabulary Object Detection (OVOD)~\cite{gu2021open, zhou2022detecting}.
 These detectors can be  broadly considered as two-stage detectors~\cite{gu2021open,zhong2021regionclip,du2022learning,feng2022promptdet,kuo2022f} 
and one-stage detectors~\cite{xie2021zsd,rao2021denseclip,ma2022open}.
 Although  a one-stage  detector is inherently simpler and
    less costly to compute, it suffers from poorer performance than
    that of a two-stage detector using CLIP~\cite{ma2022open}. 
    In this work, we propose a CLIP-based one-stage detector GridLCIP that narrows the
    performance gap from typical two-stage detectors,
    while requires a much shorter training time (43$\times$ less compared to ViLD~\cite{gu2021open}) and test time (5$\times$ faster) .

Specifically, we \jy{learn} 
    grid-level representations of images that can be further used for classification, since an object in one-stage detectors is noted by the category of a grid (a pixel in a feature map) and its corresponding bounding box.
Therefore, we expand the
    conventional CLIP image-text holistic mapping to grid-text
    mapping, namely \emph{Grid-level Alignment}. 
Although some other one-stage detectors~\cite{ma2022open} 
    also share similar spirits, 
    they align the image embeddings trained 
    from supervised or visual data only self-supervised learning methods
    (rather than CLIP-like visual-language pretrained models) to align with CLIP text embeddings. 
While we directly adapt the CLIP image embeddings to generate grid-level embeddings, 
which directly benefits from the generalizability of the CLIP image encoder 
and is intrinsically more consistent with CLIP text embeddings.
 However, similar to DenseCLIP~\cite{rao2021denseclip} 
    which can only perform close-set detection, 
     grid-level alignment is only performed on the base categories 
    without  the scope to learn knowledge of novel  (unseen) categories
     for open-set detection.
 
To further exploit CLIP to learn representation for novel categories, 
    some approaches~\cite{ zhou2022detecting, ma2022open,
      gao2021towards, feng2022promptdet}  used extra
    image-caption  or  labeled datasets like
    CC3M~\cite{sharma2018conceptual} or
    ImageNet-21K~\cite{deng2009imagenet},  
    making the training process complicated and resource-consuming. 
 In this work, we want to explore CLIP directly without the need
  of extra image-caption
  and/or labelled training data in order to
    learn novel categories by applying knowledge distillation on
    CLIP.
As revealed in HierKD~\cite{ma2022open}, 
    the gap between one-stage and two-stage detectors is that the knowledge distillation on 
    two-stage detectors happen on image regions of both base and novel categories~\cite{gu2021open,zhou2022detecting} 
    given by a  separate pretrained region proposal network  (RPN), hence
    two-stage, while the one-stage detectors mainly only use base categories for knowledge distillation~\cite{xie2021zsd}.
Therefore, HierKD aligns the text embedding of the caption 
    to its paired image embedding,
    and further 
    uses an attention layer for adapting the gap between captions and images. 
However,  HierKD requires the training images to have paired
captions in addition to detection annotations, making it
  unscalable to training on other detection datasets for more general detection
  tasks. 
 To avoid all these additional requirements on model training, we
  propose to use visual-to-visual alignment instead of
   caption-to-visual alignment. This is designed
    to learn the visual embeddings of both base and novel categories
     without the need for further adaptation.
We call it \emph{Image-level Alignment}.
Specifically, we align the image-level embeddings with the one that is generated by a fixed CLIP image encoder (teacher).
Since the feature extractor of image-level embeddings and grid-level embeddings are mainly shared, the grid-level embeddings can implicitly obtain knowledge 
of undersampled categories 
from the teacher. 
In this way, we \jy{implicitly train} the grid-level embeddings 
to align to both base and novel categories in CLIP space.

%
 Overall, we propose a one-stage \jy{detector}  
    GridCLIP,
     which exploits CLIP to supplement the knowledge of undersampled categories in downstream detection datasets by simultaneously applying 
    grid-level and image-level alignments.
Our contributions are:
 (1) We exploit CLIP to
supplement the missing knowledge of undersampled object
  detection categories  in training a \jy{one-stage} detector, mitigating
the poor performance due to the long-tail data distribution in most
existing detection training data.
 (2) We propose a simple yet effective visual-to-visual knowledge
  distillation method for learning novel categories in
    constructing a one-stage CLIP-based detector, providing
  2.4 AP gains  on novel categories  compared to the  baseline.
(3)
\jy{GridCLIP is} capable of 
   handling Open-Vocabulary Object Detection with considerable
  scalability and generalizability,
 reaching the
    comparable performance to two-stage detectors in Open-Vocabulary Object Detection with much higher training and inference speed,
   without using extra pretraining processes or additional
     fine-tuning datasets.

\section{Related Works}


\noindent \textbf{Vision-Language Pretrained Model (VLM).}
Visual-only pretrained model has dominated the pretraining process
for several years until VLMs have appeared.
In comparison, 
Vision-Language Pretrained Models (VLMs) are able to 
align more visual concepts out of manual predefined categories
to natural language, extending the generality of the model.
Recently, 
a considerable number of vision-language pretrained models~\cite{radford2021learning,singh2021flava,jia2021scaling,yuan2021florence} emerge,
training from large-scale image-text data 
in an unsupervised way.
These models usually have both image and text encoders to generate corresponding features that can be aligned in a cross-modality representational space for corresponding image-text matching.
Utilizing these alignment spaces helps
zero-shot transfer 
to a wide range of downstream visual recognition tasks, such as
object detection~\cite{huang2022unsupervised,du2022learning,feng2022promptdet}, 
segmentation~\cite{zhou2021denseclip,luddecke2021prompt}, 
image retrieval~\cite{zhang2021vinvl,li2021align}.
As one widely-used instance, 
CLIP~\cite{radford2021learning} is trained on 400 million image-text pairs 
through a contrastive objective,  which extends significantly the
        generalizability and usability of
        the learned image 
        embeddings to 
        align to broad categories,
showing competitive performance 
with its fully supervised counterparts. 
CLIP is widely applied 
both in downstream tasks oriented pretraining~\cite{zhong2021regionclip,li2022grounded} 
and in fine-tuning for downstream tasks~\cite{gu2021open,rao2021denseclip}.



\noindent \textbf{Object Detection using VLM.}
OVR-CNN~\cite{zareian2021open} is the first to utilize natural language (captions) for object detection. 
\jy{While recent detectors apply large-scale image-text datasets
to learn generalizable image representations. 
Some VLMs like GLIP~\cite{li2022grounded} and DetCLIP~\cite{zhang2022glipv2} utilize large-scale annotation datasets in addition to image-text pairs for pertaining, which requires relatively high annotation cost. 
While we focus on utilizing unsupervised VLMs like CLIP and learn from annotation datasets of a limited number of categories to transfer to broader categories.
 }

To learn knowledge of base categories, 
    most detectors~\cite{gu2021open,zhou2022detecting} replace the classifiers of their detection heads with VLM text embeddings.
    Recent detectors mainly improve the learning in two aspects: learning better text embeddings of categories and extracting image embeddings to align with these text embeddings. 
(1) For generating better text embeddings, 
    also called \emph{Prompt Learning},
current approaches can be roughly classified as template-based and learnable prompting. 
The template-based one uses fixed incomplete sentences 
    that can accept labels to build complete sentences~\cite{gu2021open,ma2022open,zhong2021regionclip},
    while the learnable prompting methods concatenate learnable parameters with category labels as the input, 
    where the prompt is implicitly learned during fine-tuning~\cite{zhou2021learning,du2022learning,feng2022promptdet}.
GridCLIP uses template-based prompting as in the original CLIP and some OVOD detectors~\cite{gu2021open,ma2022open,zhong2021regionclip} for simplicity and scalability.
%
(2) For extracting image embeddings to align with text embeddings, two-stage
    detectors~\cite{ren2015faster,sun2021sparse,gu2021open, zhou2022detecting, du2022learning,zhong2021regionclip} use the embedding of cropped object bounding-box proposals.
However, they need to train a region proposal network first and require multiple inferences of the CLIP image encoder to compute the visual embedding for each region, 
    which is relatively inefficient. 
In comparison, a one-stage
    detector~\cite{lin2017focal,tian2019fcos, rao2021denseclip} aligns parts
    in an image 
    represented by grid-level embeddings. 
DenseCLIP~\cite{rao2021denseclip} 
    adapt it as
    aligning grid-level image features with text, 
    while can only perform under close-set settings. 
 While HierKD~\cite{ma2022open} performs image-level, region-level and grid-level alignment to CLIP, 
    which however requires the match captions with the detection dataset, 
    limiting its deployment to other downstream detection datasets.
Our method uses grid-level alignment for efficiency 
    while preserving the original alignment space of CLIP to gain better generalization ability, 
     without requiring extra datasets.

To learn knowledge of novel categories, some approaches~\cite{ zhou2022detecting, ma2022open, gao2021towards, feng2022promptdet} use extract knowledge from external datasets extra image-caption or labeled datasets like CC3M~\cite{sharma2018conceptual} or ImageNet-21K~\cite{deng2009imagenet}, 
    making the training process complicated and resource-consuming.
While we argue that CLIP has been trained over a broad vocabulary and 
     has the ability to provide visual embeddings of various categories. 
Therefore, we explore the original CLIP representation space 
    to learn novel categories by applying knowledge distillation on CLIP as in ViLD~\cite{gu2021open}. 
    %
 

\begin{figure*}[t]
   \centering
   \includegraphics[width=16cm]{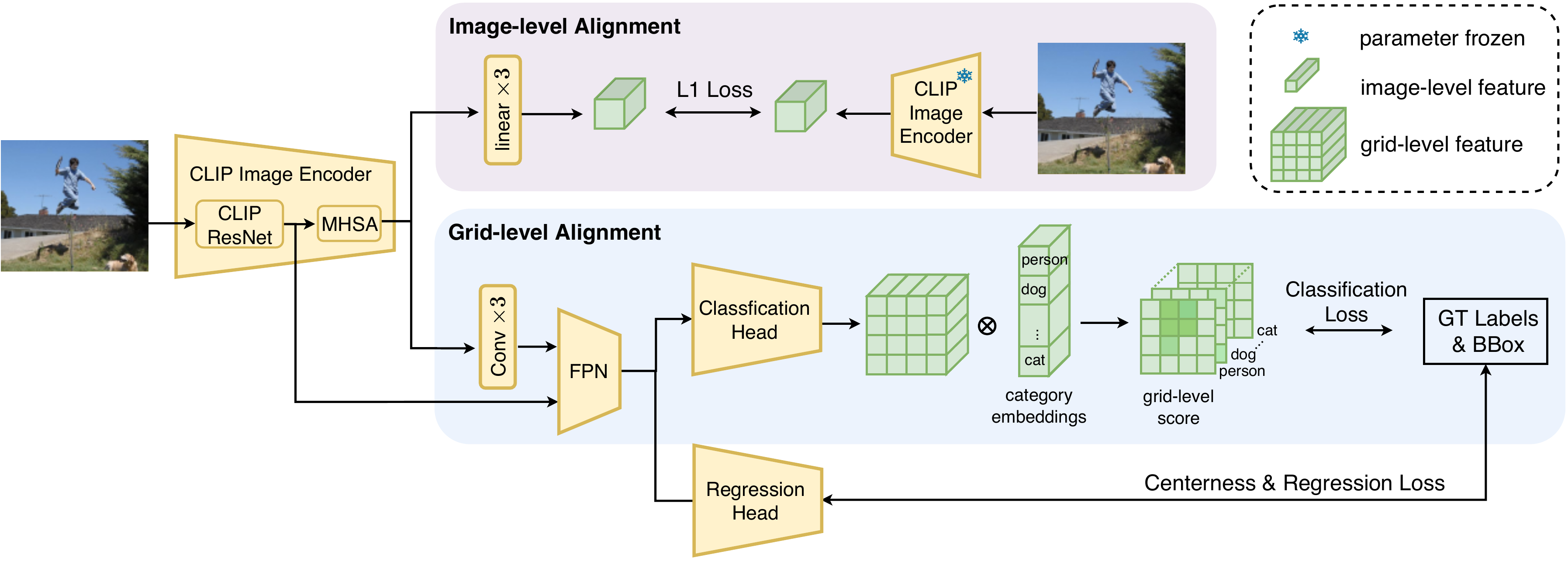}
   \hspace{10pt}
   \caption{The pipeline of our proposed GridCLIP. GridCLIP aligns to the CLIP representation in both image and grid levels. In image-level alignment, the image-level feature is aligned to the feature generated by a fixed CLIP image encoder. In grid-level alignment, the grid-level feature is aligned to the classification target generated from the ground truth labels and bounding boxes following the detector FCOS~\cite{tian2019fcos}. Note that the grid-level alignment is performed in multiple scales, while only one scale is presented here for simplicity.    \label{framework}}
\end{figure*}

\section{Approach}

The overall model design of GridCLIP is shown in
    Figure~\ref{framework}. 
We first introduce
    the strategy of adopting the CLIP embeddings for a detection task, and
    then present the approach to mapping simultaneously the CLIP
    representation by both grid-level and image-level alignments
    based on a one-stage detector FCOS~\cite{tian2019fcos}. 

\subsection{Adapting CLIP for Detection}\label{sect3_1}
CLIP consists of an image encoder (ResNet~\cite{he2016deep} or
    ViT~\cite{dosovitskiy2020image}) 
    and a text encoder (Transformer~\cite{vaswani2017attention}), 
    which together form
    the alignment space of visual and language embeddings. 
However, as the image
    embedding in CLIP is a high-dimensional feature vector of an entire image
    instead of a spatial region or pixels,
    and the text embedding is encoded from a
    sentence  instead of  a single category label in detection, 
    further adaptation for the detection task is needed. 

\noindent \textbf{Generating Image Embedding.}
The original CLIP image feature ${\overline{z}}$  is a single
    high-dimensional feature vector representing an entire image
    without spatial information.  
To get the grid-level feature ${z}$ , inspired by DenseCLIP~\cite{rao2021denseclip}, we use the other
feature from the last layer 
of the CLIP image encoder. 
Specifically,
taking the ResNet50 encoder as an example, the final output feature in the 5-th stage ${C}_{5} \in \mathbf{R}^{H_{5}  \times W_{5} \times D_{5}}$ is  
 first performed global average pooling to
 get the image-level feature $\overline{{C}}_{5} \in \mathbf{R}^{1 \times 1 \times D_{5}}$ , 
where $H_5$, $W_5$, $D_{5}$ are the height, width and number of channels of the feature in the 5-th stage of the ResNet50.
Then the concatenated features $\left[\overline{{C}}_{5}, {C}_{5}\right]$ 
are fed into a Multi-Head Self-Attention (MHSA) layer~\cite{vaswani2017attention} as follows,
\begin{eqnarray}
   [\overline{{z}}, {z}] ={ \rm{MHSA}}\left(\left[ \overline{{C}}_{5}, {C}_{5}\right]\right).
\end{eqnarray}

In CLIP, the output with spatial information ${z}$ is dumped and $\overline{{z}}$ is used to match with the text embedding. 
However, as illustrated in DenseCLIP, since the MHSA is symmetric to each input element, ${z}$ may behave similarly 
to $\overline{{z}}$, which aligns well with the text embedding. 
Therefore, we adopt both ${z}$ and $\overline{{z}}$ to generate our grid-level and image-level embeddings respectively with a few adaptation layers.



\noindent \textbf{From Label to Text Embedding.}
In the original CLIP,
to create a dataset classifier from label text,
a set of template-based prompts 
like ``a photo of a \{\emph{object}\}.'' are applied, 
where \emph{object} is any of the target category names. 
Then the multiple prompts for a single label are aggregated.
Although there are several learnable prompting methods for CLIP~\cite{zhou2021learning} that fine-tune with the downstream tasks, 
We follow the template-based  one for simplicity and scalability.
We use one template-based  variant in ViLD~\cite{gu2021open} designed for object detection.
Here we note the final text embeddings of the categories in the target dataset ( base categories) as 
$\left\{T_{k}\right\}_{k=1}^{K}$, where $K$ is the number of category.






\subsection{Grid-level Alignment}
\noindent \textbf{Generating Grid-level Image Embedding.} Taking ResNet50 encoder as an example, in FCOS~\cite{tian2019fcos}, the output feature map {${C}_{3}$, ${C}_{4}$, ${C}_{5}$} of ResNet50 are inputted into FPN, 
producing $5$ multi-scale image feature maps $\left\{{P}_{i}\right\}_{i=3}^{7}$. In FPN, ${C}_{5}$ is fused with ${C}_{3}$, ${C}_{4}$ to produce ${P}_{3}$, ${P}_{4}$ as well as serves as the input of ${P}_{6}$, ${P}_{7}$.
Therefore, we fuse ${z'}$ into ${C}_{5}$ to spread the image embeddings suitable for the text alignment to different scale image feature maps.

To be specific, we first apply three consecutive 3$\times$3 convolutional layers with ReLU activation function to adapt the MHSA grid-level output feature $z$, reducing the number of channels from 1024 to 256 to generate $z'$, and concatenate it with ${C}_{5}$.
Then the concatenated feature $\left[{{z'}}, {C}_{5}\right]$ replaces ${C}_{5}$ and is fed into the FPN with a little modification in the input channel number. 
In this way, the FPN is able to produce the multi-scale feature maps $\left\{{P}_{i}\right\}_{i=3}^{7}$ inheriting from the CLIP image embeddings that can be aligned to the text embedding as formulated below,
\begin{eqnarray}
   \left\{{P}_{i}\right\}_{i=3}^{7}={\rm{FPN}} ({C}_{3}, {C}_{4}, \left[{{z'}}, {C}_{5}\right]).
\end{eqnarray}

The FPN output feature $\left\{{P}_{i}\right\}_{i=3}^{7}$ are then used to generate the final multi-scale grid-level features $\left\{{G}_{i}\right\}_{i=3}^{7}$
by going through the FCOS classification head.
In the original FCOS,  the classification head contains 5 convolutional layers and the last layer outputs the features with the channel number equal to the category number.
While in GridCLIP, 
We instead modify the output channel to be equal to the dimension of the CLIP text embeddings, to generate $\left\{{G}_{i}\right\}_{i=3}^{7}$.
Then for each scale, 
    the output feature map calculates the cosine similarities 
    with each text embedding (each category) 
    in pixel level 
    corresponding to grids in the original image,
    to produce the multi-scale grid-level score with the Sigmoid activation function.
For any grid (pixel) $j$ in the $i$-th scale grid-level feature $G_i(j)$, the matching score over all categories can be formulated as below,
 \begin{eqnarray}
   S_i(j) = \left\{  \dfrac{ G_i(j)\cdot T_k}{{\left\| G_i(j)\right\| _{2}\left\| T_k\right\| _{2}} }\right\}_{k=1}^{K}.
\end{eqnarray}

\noindent Finally, the grid-level score  $\left\{{S}_{i}\right\}_{i=3}^{7}$ is treated as the original classification output and aligned to the ground-truth target  $\left\{{Target}_{i}\right\}_{i=3}^{7}$  using Focal Loss~\cite{lin2017focal} as in the original FCOS. 

\subsection{Image-level Alignment}
With grid-level alignment, the image 
 grids
of the  base categories are mapped to the CLIP alignment space, 
by aligning their embeddings to the corresponding text embeddings $\left\{T_{k}\right\}_{k=1}^{K}$.
While for 
 grids of  novel categories, 
    there are no corresponding text embeddings to align to,
    which can only learn their embeddings by minimizing their similarity to any of $\left\{T_{k}\right\}_{k=1}^{K}$ during training.
However, since the embeddings of different  novel categories 
    can have different similarities to each  base category, 
    simply minimizing the similarities
    between the  base and  novel categories
    is not consistent with the CLIP representation space 
    which presents a generalizable knowledge representation.
Therefore, ignoring the alignment of  novel categories
    may limit the ability to encode a wide range of  novel visual concepts, which
    harms the generalization ability of the model.

In practice, inspired by ViLD~\cite{gu2021open}, 
    we align the image-level embedding ${\overline{z}}$
    to the embedding ${\overline{z}_{ \rm CLIP}}$ produced by a fixed CLIP image encoder, 
    so that the regions of  novel categories in an image can also
    be projected to the CLIP alignment space.
Different from ViLD that 
    aligns the embedding of several proposed regions in an image
     provided by a separate region proposal network (RPN), hence two-stage,
    which is the source of significant extra computational costs due to
    its requirement for multiple inferences of the image encoder for each
    object proposal, we directly align the embedding of the whole
    image ${\overline{z}'}$  without the need for multiple passes.
Specifically, 
    similar to  grid-level alignment, we generate the image-level embedding $\overline{{z}'}$
    from $\overline{{z}}$ going through three consecutive linear layers with the ReLU activation function.
    Then we maximize the $L_1$ similarity between ${\overline{z}}$ and ${\overline{z}_{ \rm CLIP}}$ , the reverse of which is the image-level alignment loss  $L_{ \rm image}$.
As for the fixed CLIP image encoder, we evaluate different published versions of pretrained models
    and choose the ViT-B/32 version for alignment. 
Note that image-level alignment is only performed during the training phase.

Finally, the total loss for training GridCLIP end-to-end is,
\begin{eqnarray}
  L = w_{ \rm grid} L_{ \rm grid} + w_{ \rm image} L_{ \rm image} + L_{ \rm R} + L_{ \rm C}, 
\end{eqnarray}
\noindent which includes the loss of two alignments as well as the original loss in the one-stage detector FCOS: Regression loss $L_{ \rm R}$ for bounding boxes  and  centerness loss $L_{ \rm C}$ indicating the distance of a pixel to the center of the bounding box.

\section{Experiments}

\setlength{\tabcolsep}{8pt}
\begin{table*}[]
   \centering
   \footnotesize
   \caption{Comparison with different object detectors on LVIS v1.0~\cite{gupta2019lvis}  with open-set settings. Multi-scale training is used. ``CLIP on cropped regions'' directly applies CLIP to classify cropped region proposals. 
   \jy{Except for GridCLIP, all detectors use an RPN pretrained on base categories to get region proposals.
   $\ddagger$ denotes using mask annotations.
   $\dagger$ denotes mask AP. 
   $\nabla$ denotes using learnable prompts instead of template prompts.
   $\star$ denotes that RegionCLIP use extra pretraining process of 600k iter on CC3M dataset with batch size of 96. 
   }
}\label{sota_lvis_open}
\begin{tabular}{l|cccc|c|cc|c}
\toprule
 Model   & backbone   & pretrained CLIP    & 
 epochs & 
 {\begin{tabular}[c]{@{}c@{}}external\\ dataset\end{tabular}}    & AP$_{r}$ & AP$_{c}$ & AP$_{f}$ & AP   \\ \hline
{CLIP on cropped regions}$\ddagger$~\cite{gu2021open} & {R50-FPN  } & {ViT-B/32} & 0      &    { }   &   \textbf{19.5}     & 19.7     & 17.0  & 18.6 \\
ViLD$\ddagger$~\cite{gu2021open}   & {R50-FPN  } & {ViT-B/32} & {384}  &    { } &   16.3     & 21.2     & 31.6     & 24.4 \\ 
RegionCLIP~\cite{zhong2021regionclip} & CLIP R50-C4 & RN50  & 12$\star$ & CC3M &  17.1     & \textbf{27.4}     & \textbf{34.0}     & \textbf{28.2} \\
Detic$\ddagger$$\dagger$~\cite{zhou2022detecting} & R50-FPN  & ViT-B/32     & 384    & ImageNet-21K & 17.8    &  \underline{26.3}    & 31.6     & \underline{26.8} \\  
PromptDet$\nabla$$\dagger$~\cite{feng2022promptdet}  & R50-FPN    & ViT-B/32 & 6+12 & LAION-novel  & \underline{19.0}& 18.5     & 25.8     & 21.4 \\
\hline

GridCLIP-R50   & {CLIP  R50-FPN}  & ViT-B/32 &   24  &    &15.0 &	22.7&	32.5&	25.2 \\
GridCLIP-R50-RN   & {CLIP  R50-FPN}  & RN50x64  &   24  &    &13.7	&23.3	&\underline{32.6}	&25.3 
\\

\bottomrule
\end{tabular}

\end{table*}

\setlength{\tabcolsep}{10pt}

\begin{table*}
   \centering
   
   \footnotesize
   \caption{\jy{The training and test time on LVIS v1.0 and the model size comparisons of ViLD and GridCLIP. The resource usage of ViLD is measured based on the implementation of DetPro~\cite{du2022learning}.  }} \label{resource}
   
   \begin{tabular}{l|c|ccc|c}
   \toprule
             Model & \begin{tabular}[c]{@{}c@{}}Parameters \\      for Inference (M)\end{tabular} & Epoch 
             & \begin{tabular}[c]{@{}c@{}}Training Cost     / Epoch \\ (Per-GPU-Hour)\end{tabular} 
             & \begin{tabular}[c]{@{}c@{}}Total Training Cost   \\ (Per-GPU-Hour)\end{tabular} & FPS   \\\hline
ViLD        & 60.5 &384      &7.98&	3064      & 3.3               \\ \hline
GridCLIP-R50 & 56.4   &24    &2.94&	70               & 19.5                       \\
GridCLIP-R50-RN & 56.4  &24  &3.66&	88                   & 19.5                       \\
\bottomrule
\end{tabular}
\end{table*}

\setlength{\tabcolsep}{8pt}
\begin{table}[]
   \centering
    \footnotesize
   \caption{Generalization ability of LVIS-trained detectors to PASCAL VOC 2007 test set and COCO validation set.}\label{transfer_lvis_open}

\begin{tabular}{l|cc|ccc}
\toprule
 &  \multicolumn{2}{c|}{PASCAL VOC}  & \multicolumn{3}{c}{COCO} \\
  \multirow{-2}{*}{Model} & AP$_{50}$    & AP$_{75}$    & AP   & AP$_{50}$    & AP$_{75}$\\ \hline
ViLD & {72.2}    & {56.7}    & 36.6    & {55.6}     & {39.8}    \\
DetPro & {74.6}    & {57.9}    & 34.9    & {53.8}     & {37.4}  \\ \hline

GridCLIP-R50    &70.9	& 55.4 &	34.7&	52.2&	37.1\\
GridCLIP-R50-RN    &71.6&	55.7&	34.4&	51.8&	36.6\\
\bottomrule
\end{tabular}
\end{table}

\subsection{Implementational Details}


GridCLIP uses the one-stage detector FCOS~\cite{tian2019fcos} as the detector, which can be replaced by other one-stage detectors like RetinaNet~\cite{lin2017focal} or ATSS~\cite{Zhang_2020_CVPR}.
The backbone uses the RN50 pretrained CLIP image encoder, which has two more convolutional layers than the original ResNet50~\cite{he2016deep} in the stem module 
    and a Multi-Head Self-Attention (MHSA) layer performing on the output of the 5th stage.
The adapting layers performed on the output features of MHSA use the embedding dimension of 256.
For image-level alignment, GridCLIP uses the ViT-B/32 version of CLIP.  
The weights of the two alignment losses are: $w_{ \rm grid}$=1, $w_{ \rm image}$=10.
Our implementation is based on the MMDetection framework~\cite{chen2019mmdetection}.

We conduct experiments on the detection benchmark 
 LVIS v1.0~\cite{gupta2019lvis}.
LVIS v1.0 is a long-tail detection dataset containing 1203 categories.
 The categories are divided into three parts by how many images
they appear in: rare (1-10), common (11-100), and
frequent ($>$100), respectively including 337, 461 and 405 categories, with corresponding metrics AP$_r$,  AP$_c$ and AP$_f$.
Following the ViLD~\cite{gu2021open}, we use frequent and common categories as the base categories and rare categories as the novel categories 
for open-set detection. For close-set detection, we use the common and frequent categories.
When comparing with other SOTA approaches, we adopt multi-scale training similar to ViLD and random cropping augmentation. 
For the training process,
     GridCLIP is trained for a 2$\times$ (24 LVIS epochs) schedule with a batch size of 16.
During the inference stage, the maximum number of detection objects per image is 300,  and the threshold of the classification score is set to 0.05.
The IOU threshold of NMS is 0.5. 
Refer to the supplementary materials for more details. 

\subsection{Comparison with the State-of-the-Art}\label{sec_sota}

We compare GridCLIP on the LVIS v1.0 validation set with other methods with comparable backbone, including ViLD~\cite{gu2021open}, RegionCLIP~\cite{zhong2021regionclip}, Detic~\cite{zhou2022detecting},  DetPro~\cite{du2022learning} and PromptDet~\cite{feng2022promptdet} in Table~\ref{sota_lvis_open}.
\jy{These methods use template-based prompts, 
except that DetPro~\cite{du2022learning} and PromptDet~\cite{feng2022promptdet} use learnable prompts.}
Also, we do not compare to methods like GLIP~\cite{li2022grounded} and DetCLIP~\cite{zhang2022glipv2} which use large-scale annotation data, since we focus on utilizing limited annotation data for the detection of broader categories.

 \noindent \textbf{Performance Comparison.}
A totally fair comparison is not realistic, since external datasets or learnable prompts are widely used in most OVOD methods.
Therefore, we find it relatively fair to compare GridCLIP with ViLD~\cite{gu2021open}
which only utilizes the knowledge of CLIP and the detection dataset without learnable prompts. 
We observe that GridCLIP surpasses ViLD in overall AP by 0.8. 
As a one-stage detector, 
GridCLIP closes the gap to the two-stage detector ViLD in novel categories to 1.3 AP$_{r}$, 
    while the current SOTA one-stage detector HierKD is still 8.2 AP behind ViLD on the COCO validation dataset.
Furthermore, GridCLIP outperforms ViLD by 1.5 AP$_{c}$ and 0.9 AP$_{f}$.
Besides ViT-B/32, we also use the RN50$\times$64 version (the largest model of CLIP under ResNet architecture) of CLIP for image-level alignment
    to explore the upper bound of the ResNet version. We observe that the RN50$\times$64 version has a worse generalization ability to novel categories compared to the ViT-B/32 one,
    with lower AP$_{r}$ while higher  AP$_{c}$ and AP$_{f}$.  
\jy{We further observe the obvious gap between the base and novel categories in GridCLIP, 
and try to understand and explain the gap based on the analysis from another one-stage detector HierKD~\cite{ma2022open}.
In ViLD, 
    both the novel and base categories \jy{use} the region-level alignment.
In GridCLIP,
    \sgg{whilst} novel categories \sgg{use} the more coarse-grained image-level alignment, 
    \sgg{the} base categories \sgg{in GridCLIP} use \sgg{{\em both}}
    the more fine-grained grid-level alignment \sgg{and the}
    image-level alignment. \sgg{This} makes the gap between base and novel categories larger than that of ViLD. 
To verify this, we can replace image-level alignment with region-level alignment similar to ViLD to further improve AP$_{r}$, 
    which however may require more training time as other two-stage detectors do. We leave it for future research. }

\setlength{\tabcolsep}{8pt}

\begin{table*}[h]
   \centering
   \footnotesize
   \caption{Comparison of different backbones and alignment methods on LVIS v1.0~\cite{gupta2019lvis} with close-set settings. The top section compares supervised (ImageNet~\cite{deng2009imagenet}) and unsupervised (SwAV~\cite{caron2020unsupervised}) pretrained visual models with CLIP as unsupervised visual-language pretrained model. The bottom section compares different alignment methods based on the ResNet50 version of pretrained CLIP image encoder. ``GridCLIP-R50$\ast$'' uses the CLIP ResNet50 without the MHSA layer, 
    while ``GridCLIP-R50'' uses the whole image encoder of CLIP ResNet50.}\label{abla_lvis_close}
\begin{tabular}{l|ccc|cc|c}
\toprule
Method & Backbone                & \begin{tabular}[c]{@{}c@{}}Grid-level \\ Alignment\end{tabular} & \begin{tabular}[c]{@{}c@{}}Image-level \\ Alignment\end{tabular} &  AP$_{c}$ & AP$_{f}$  & AP \\
\hline
ImageNet-R50  w/o align                     &ImageNet R50-FPN       & -         & -              &    14.6	&   26.2    &	16.6 \\
SwAV-R50  w/o align                         &SwAV R50-FPN           & -         & -              &    19.5	&   29.2    &	20.0 \\
GridCLIP-R50{$\ast$} w/o align              &CLIP R50-FPN woMHSA          & -         & -        &    20.2	&  \textbf{30.3}    &	20.7 \\\hline
GridCLIP-R50 w/o align 	                    &CLIP R50-FPN       &	        &	                 &    19.4	&   29.7    &	20.1 \\
GridCLIP-R50 w grid-align 	                &CLIP R50-FPN       &\checkmark &	                 &   \textbf{21.7}	&   30.0    &	\textbf{21.2} \\
GridCLIP-R50 w image-align 	                &CLIP R50-FPN       &	        &\checkmark          &   
19.4	&\underline{30.1}	&20.2\\
GridCLIP-R50	                            &CLIP R50-FPN       &\checkmark &\checkmark          &  
\underline{21.2}	& 30.0 &	\underline{21.0} \\
\bottomrule
\end{tabular}

\end{table*}


\begin{figure*}[t]
  \centering
  \begin{tabular}{@{\hspace{-10pt}} c @{\hspace{-30pt}}  c }
    {\includegraphics[width=1.16\columnwidth]{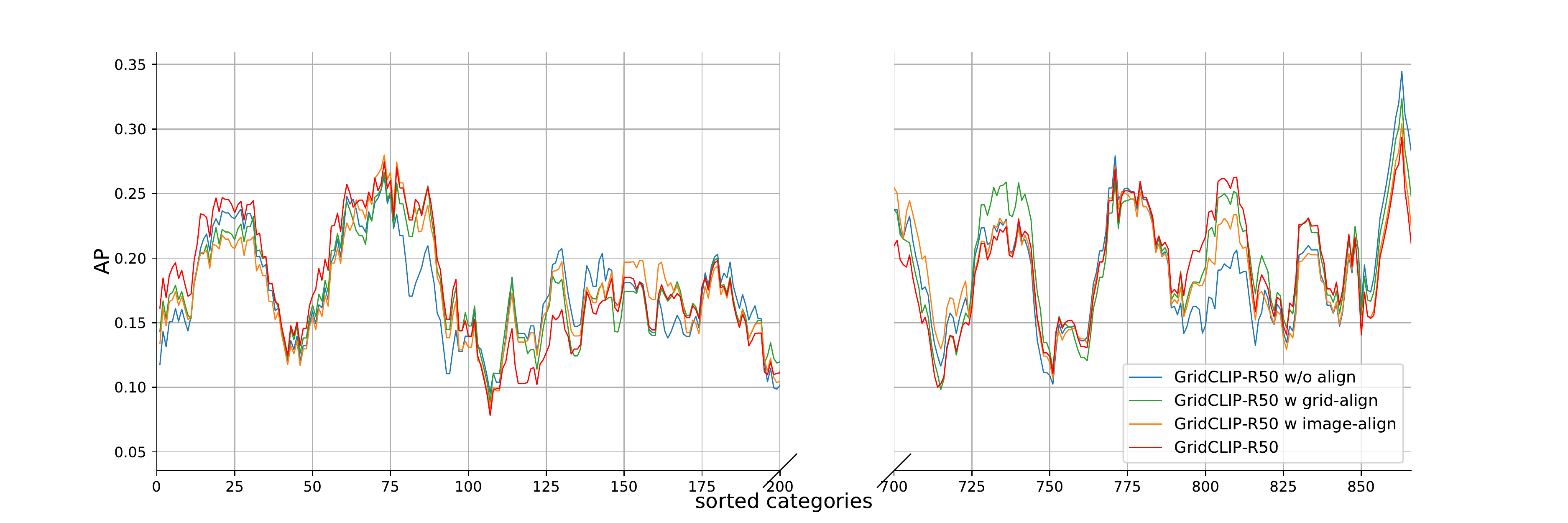}}&
      \includegraphics[width=1.16\columnwidth]{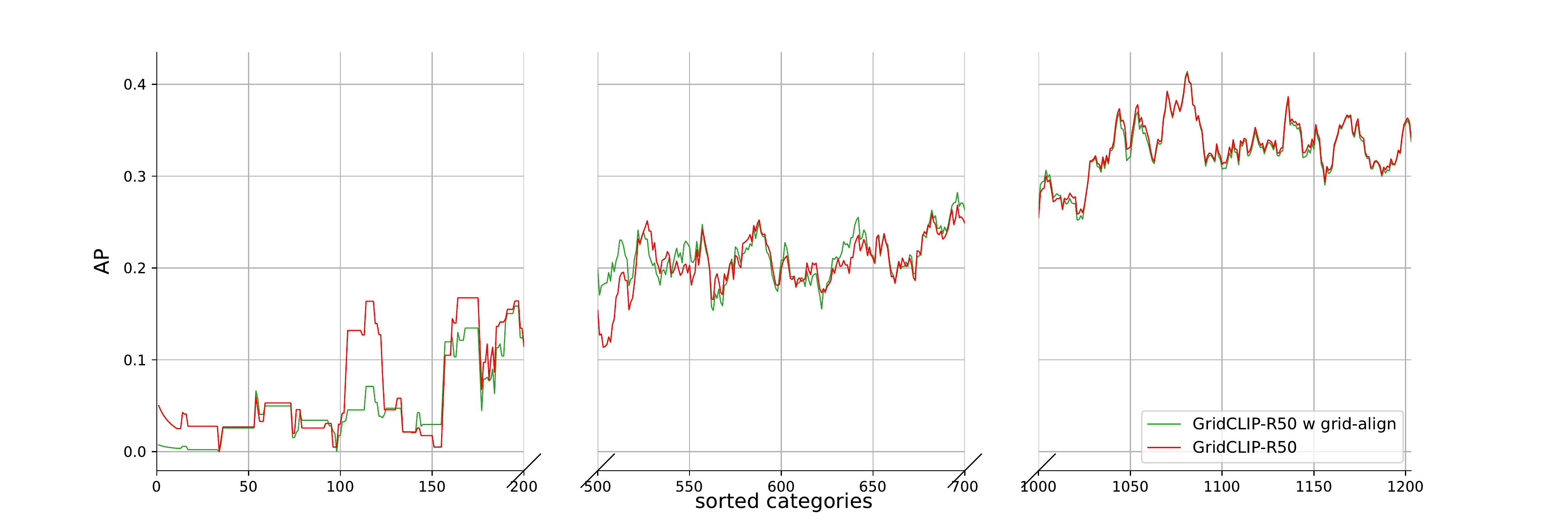} \\
    {\small (a) close-set detection} &
     { \small (b) open-set detection}
  \end{tabular}
  \caption{ The AP on LVIS v1.0 over categories sorted by frequency in ascending order.  (a) uses on the close-set setting, only containing 461 ``common''  categories and 405 ``frequent'' categories. (b) uses the open-set setting, containing containing 337 ``rare'' categories, 461 ``common''  categories and 405 ``frequent'' categories. The value is smoothed using moving average with window [-10,10].\label{abla_imbala}}
\end{figure*}


\setlength{\tabcolsep}{8pt}

\begin{table}
   \centering
   \footnotesize
   \caption{The effect of image-level alignment on LVIS v1.0~\cite{gupta2019lvis} with open-set settings.} \label{abla_lvis_open}
   \begin{tabular}{l|c|cc|c}
   \toprule
Model                        & AP$_{r}$ & AP$_{c}$ & AP$_{f}$ & AP   \\
\hline
GridCLIP-R50 w grid-align   & 10.1	         &\textbf{21.0}	&29.6	       &22.5 \\
GridCLIP-R50                &\textbf{12.7}  & 20.6  &\textbf{29.7} &\textbf{22.8}\\
\bottomrule
\end{tabular}

\end{table}

 \noindent \textbf{Resources Comparison.}
We compare GridCLIP with ViLD on training and test time, 
as well as the model size. 
ViLD is originally trained on TPUv3 with a batch size of 256. 
For fair comparison and due to resource limitation,
we train both ViLD and GridCLIP with a batch size of 16 on 2 A100 GPUs. 
We train ViLD using the implementation of DetPro~\cite{du2022learning}.
Moreover, ViLD takes 1 day on 8 V100 GPUs to pre-compute the
 CLIP image embedding of regions to accelerate training. \sgg{GridCLIP
   does not require this.}
In Table~\ref{resource}, we show that with comparable model size,
GridCLIP-R50 is approximately \sgg{43$\times$ and 5$\times$} faster than ViLD in
\sgg{training and test respectively}. 
Such significant advantages
  remain when GridCLIP-R50-RN using RN50$\times$64 (3$\times$ larger in both input and parameters) for image-level alignment, with 34$\times$ 
  faster in training time than that of ViLD. 
This validates
  clearly the compute efficiency of the one-stage GridCLIP.

 \noindent  \textbf{Transfer to Other Datasets.}
To further explore the generalizability of GridCLIP, we follow ViLD~\cite{gu2021open} and evaluate the LVIS-trained GridCLIP on \sgg{both the} PASCAL VOC 2007 test set~\cite{everingham2010pascal} and the COCO validation set~\cite{lin2014microsoft} by directly replacing the categories without any finetuning.
Note that there are overlaps of both category and image between LVIS and COCO (as well as PASCAL VOC). 
The IOU threshold
of NMS is 0.6. 
On PASCAL VOC, we observe that
the gap between GridCLIP-R50 and ViLD is 
 1.2 to 1.3 AP, 
and GridCLIP-R50-RN is comparable with ViLD on PASCAL VOC 
with no more than \jl{1 AP difference}.
Although the gap on COCO is still obvious with 2.2 AP falling behind but is comparable to that of DetPro which uses learn prompts based on ViLD.
Therefore, in the generalization ability, GridCLIP performs quite close to its two-stage counterparts.


\subsection{Ablation Studies}
We verify the effectiveness of grid-level and image-level alignment for undersampled categories 
for both close-set detection (Table~\ref{abla_lvis_close}) 
    and open-set detection (Table~\ref{abla_lvis_open}). 
We follow the same settings in Sec.~\ref{sec_sota} except 
    that multi-scaling training and random cropping augmentation
    are not used here.
\jy{Among the experiments,
``w\/o align'' denotes using the original design of FCOS only with different backbones that feed different image features to the FPN. ``w grid-align'' and ``w image-align'' denote only using grid-level or image-level alignment respectively.}

\begin{figure*}
  \centering
  \begin{tabular}{@{\hspace{5pt}} c @{\hspace{0pt}} c @{\hspace{0pt}} c @{\hspace{0pt}} c }
    {\includegraphics[width=0.52\columnwidth]{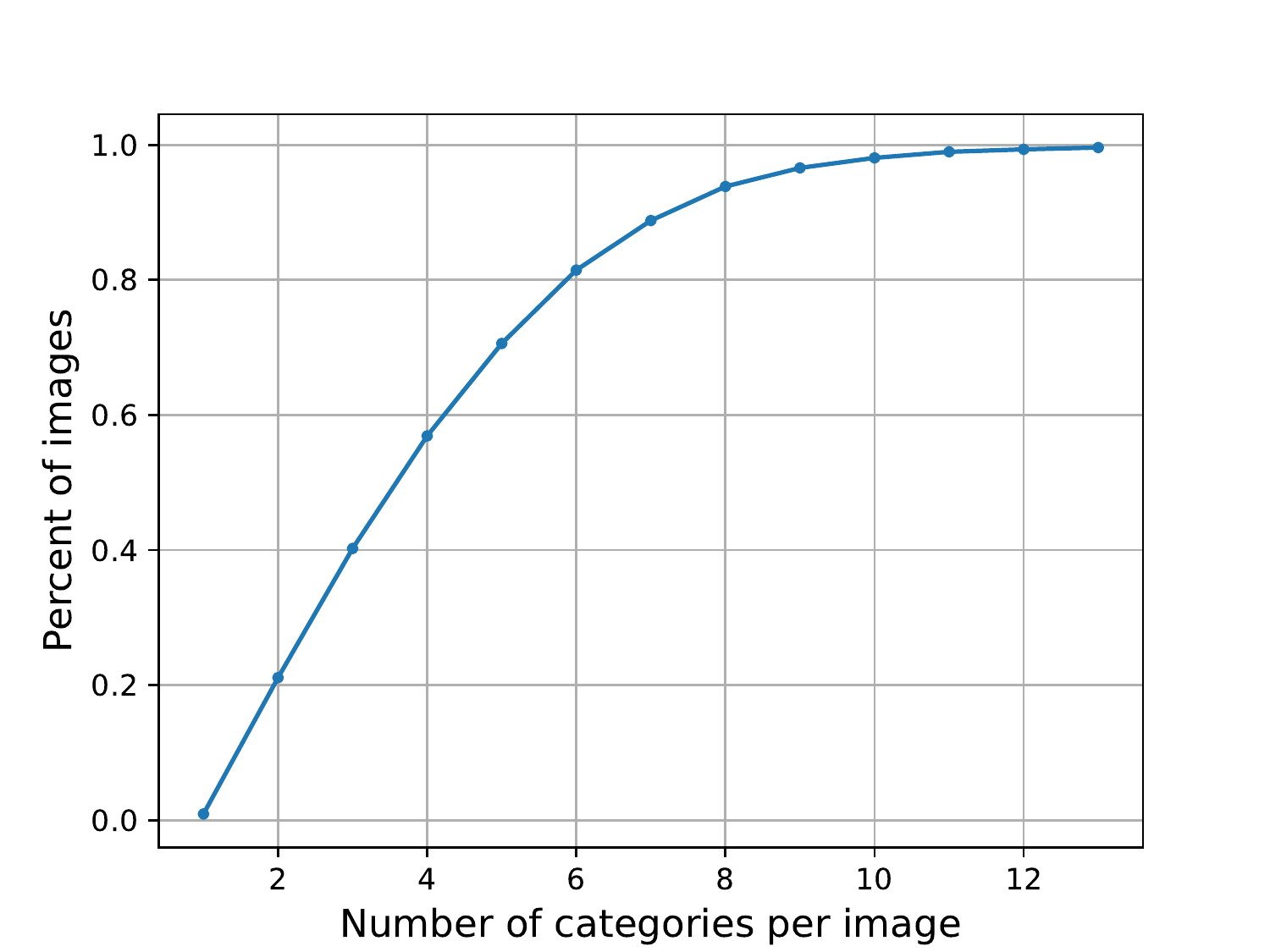}}&
      \includegraphics[width=0.52\columnwidth]{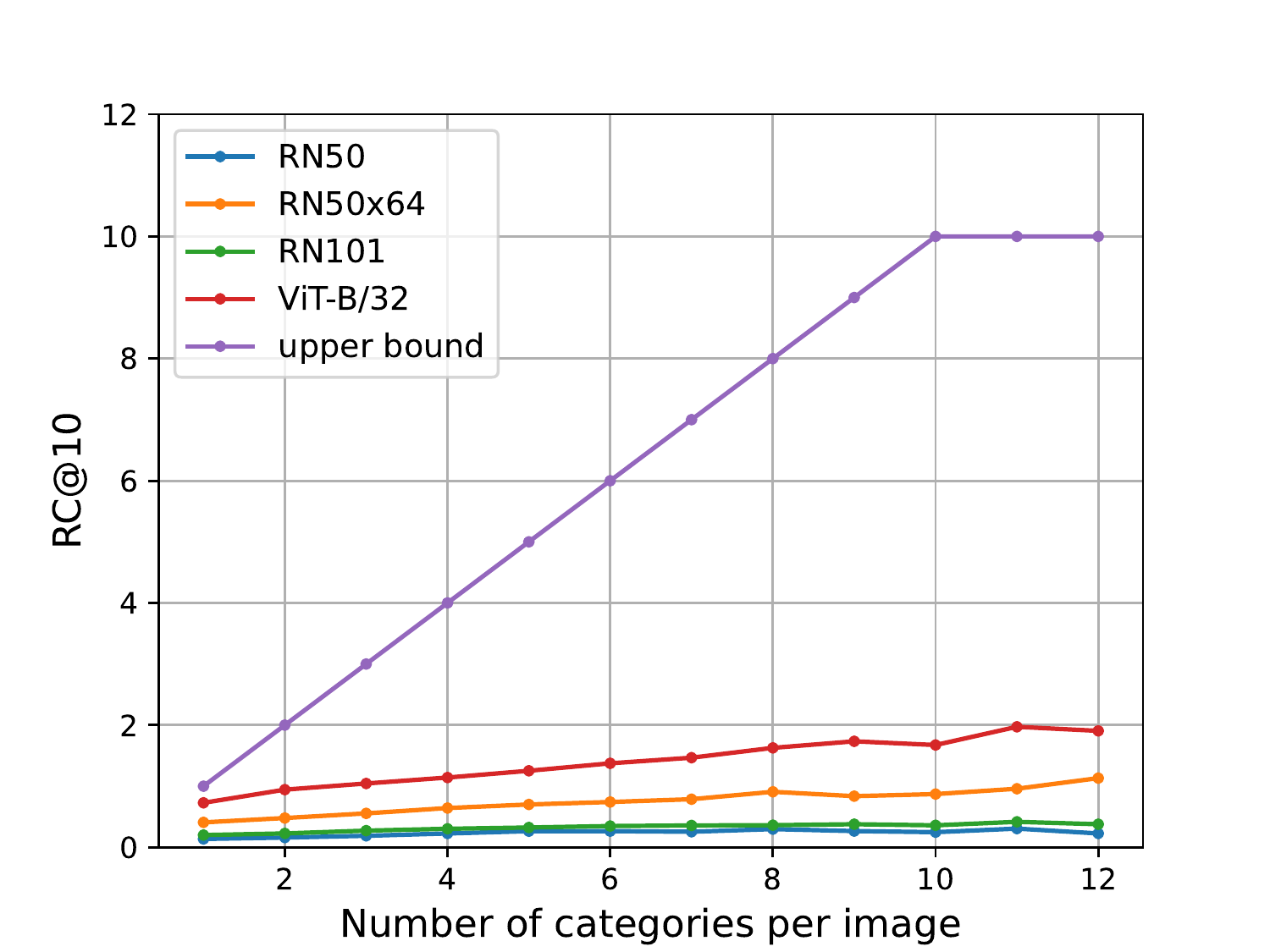} &
      \includegraphics[width=0.52\columnwidth]{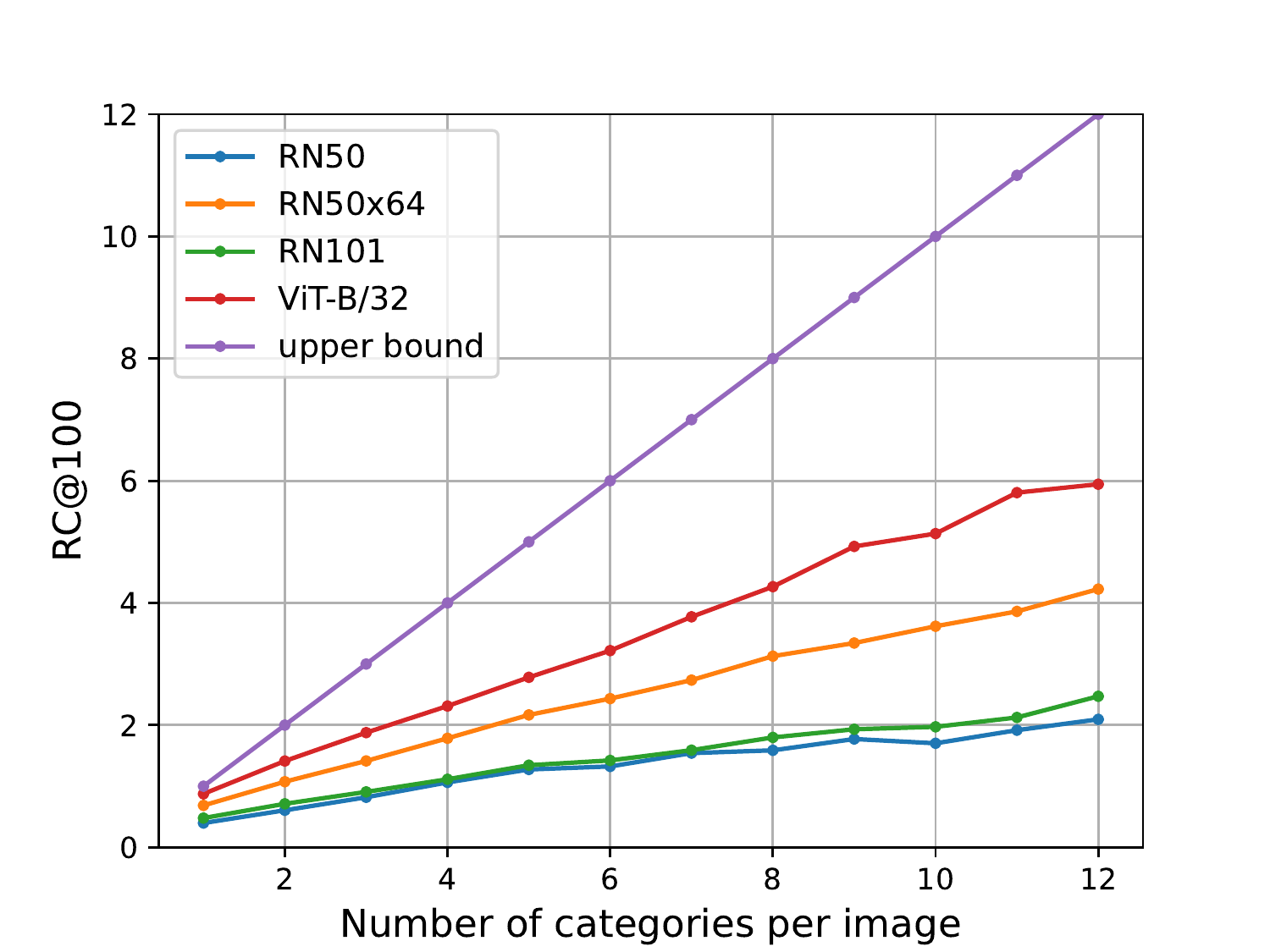}&
      \includegraphics[width=0.52\columnwidth]{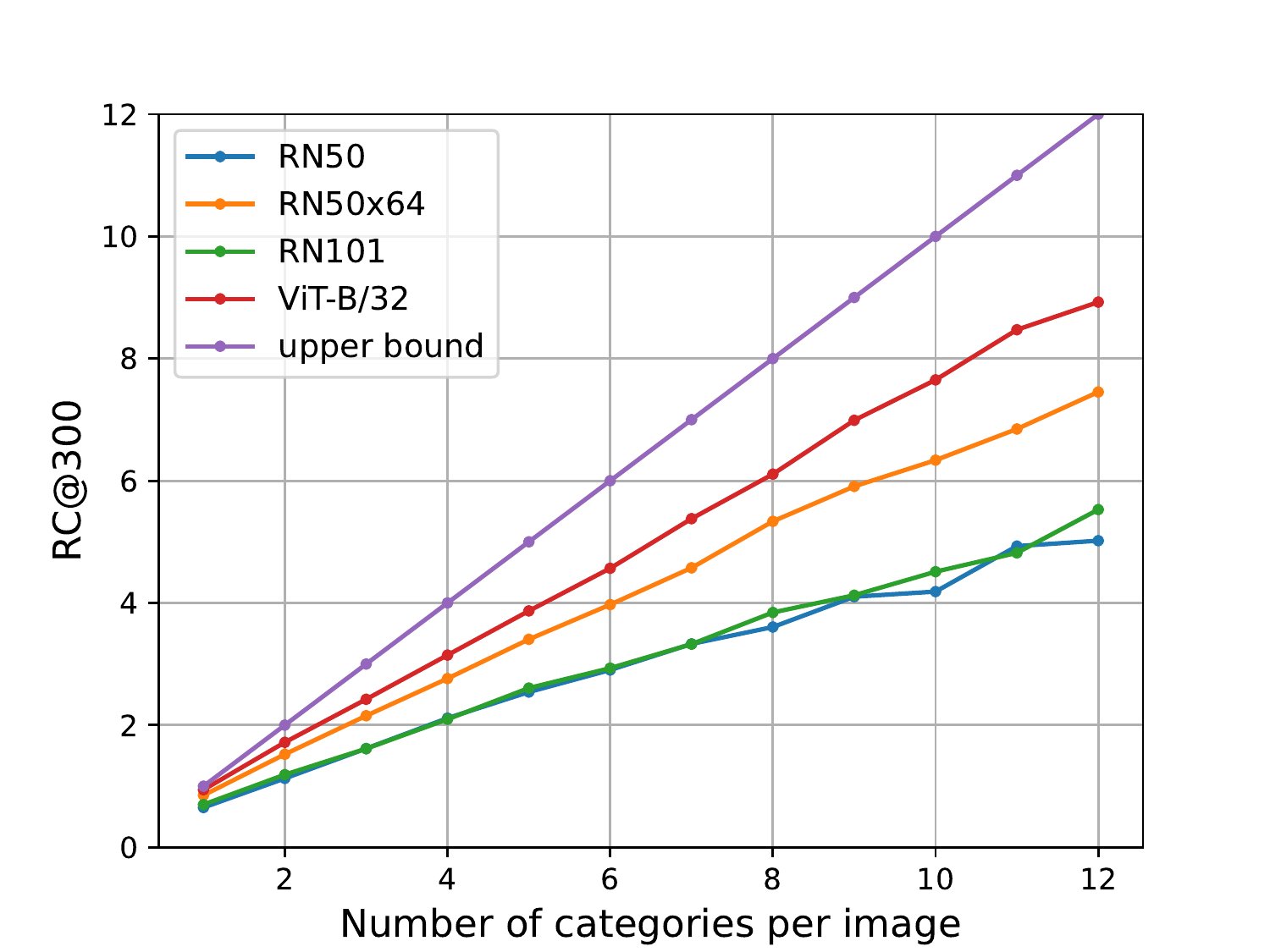} \\
    \small (a) Percent of images \label{fig:percentageClass} &
      \small (b) RC@10 &
      \small (c) RC@100 &
      \small (d) RC@300 
  \end{tabular}
  \caption{ (a) shows the percent of images containing different numbers of categories in an image on LVIS v1.0 validation set.
  (b), (c) and (d) are the recall of top k (k=10, 100, 300) predictions of different CLIP pretrained versions, using the image-level representation from the corresponding CLIP image encoder.  \label{clipHit}}
\end{figure*}

 \noindent \textbf{Close-set Detection.}
We first evaluate other visual pretrained models as the backbone to compare to the vision-language pretrained model CLIP, 
including the ImageNet~\cite{deng2009imagenet} pretrained ResNet50 on the classification task 
and self-supervised pretrained ResNet50 using visual-only SSL method SwAV~\cite{caron2020unsupervised} pretrained on large-scale unlabeled images.
By comparing the methods using different pretrained ResNet50 without any alignment (the top section of Table~\ref{abla_lvis_close}), we notice that using the CLIP pretrained ResNet50 can bring notable improvements compared to the ImageNet and SwAV pretrained ones, with the similar architecture, which indicates the superiority of the vision-language pretrained model CLIP than other visual pretrained models in generalizing better image embeddings for detection.

On the bottom section of Table~\ref{abla_lvis_close}, we introduce the MHSA layer whose output is aligned to the CLIP text encoder in the original training of CLIP. Therefore, the MHSA layer provides both grid-level and image-level features for CLIP-based alignment.
We first observe that introducing the MHSA layer without any alignment drops the overall performance by 0.7 AP,
while using both alignments can improve the performance in the infrequent common categories by 1.3 AP and preserves the performance in frequent categories.
Among the experiments that use the MHSA layer,
we find that applying grid-level alignment can significantly improve the common categories by 2.3 AP, while using
  image-level alignment seems limited on the metrics AP$_{c}$ and AP$_{f}$.
Using both alignments can bring 1.8 AP$_{c}$ and 0.3 AP$_{f}$ improvements which are lower than the one using only grid-level alignment.
We further explore the reasons behind that.
 We find that the metric of AP$_{c}$ and AP$_{f}$ are too coarse-grained for 
  distinguishing categories with different frequencies.
So we further observe the performance in a more fine-grained way by the plot of AP over categories with different sample numbers (Figure~\ref{abla_imbala}).
As shown in subfigure (a), in the 200 most infrequent categories, the improvement of applying one alignment is not stable, where the AP can be notably higher than ``GridCLIP-R50 w/o align'' in some categories while obviously worse in other categories.
By applying both alignments,
``GridCLIP-R50'' primarily ranks top in the 100 most infrequent categories, while 
its superiority is not obvious in frequent categories. 
 Therefore, we can conclude that using both alignments benefits the undersampled categories in close-set detection. 

 \noindent \textbf{Open-set Detection.}
Since only models with grid-level alignment can be extended for open-set detection by extending the categories embedding list with novel categories and using the list match with the grid-level image embeddings, we compare two models from Table~\ref{abla_lvis_close}.
As shown in Table~\ref{abla_lvis_open}, 
    image-level alignment improves novel categories by \jl{2.6} AP$_{r}$, while preserving the performance on base categories. 
Also, as shown in  Figure~\ref{abla_imbala} (b), the performance rises over categories as their training sample number increases, which verifies that undersampled categories suffer from long-tail distribution. 
In the 200 most infrequent categories which are novel (also rare) categories,  ``GridCLIP-R50'' outperforms ``GridCLIP-R50 w grid-align'' significantly, which indicates the effectiveness of  image-level alignment on novel categories. Therefore, it is verified the alignment of image-level representations also helps learn generalizable grid-level representations of undersampled categories.

    In summary, grid-level alignment improves the performance notably on undersampled categories in close-set detection and allows the detector to be extended for open-set detection, while image-level alignment obviously benefits the novel categories in open-set detection.
    Using both alignments enables a one-stage detector to detect novel categories and mitigate the deterioration of undersampled categories in long-tail datasets.

\subsection{Further Analysis: Evaluating CLIP Image-Level Representation for Object Detection}\label{evalImgCLIP}
We analyse how accurately the multiple categories in an image can be represented by the original CLIP image encoder.
This substantially affects the performance of a one-stage detector built upon the CLIP image-level representation
to detect multiple categories at the same time, which indicates how much image-level alignment can benefit GridCLIP.
We evaluate several pretrained versions of CLIP on LVIS v1.0 validation set, including the refined ResNet~\cite{he2016deep} (RN50, RN101, RN50$\times$64) and those with the Transformer architecture~\cite{dosovitskiy2020image} (ViT-B/32).
Specifically, We calculate the recall of categories by using the original CLIP image-level representation to match the text representation of each category (Figure~\ref{clipHit}).

For RC@10, all models perform poorly with no more than 2 recalls.
While for RC@100, we find that ViT-B/32 can recall 50\% of the categories and RN50$\times$64 can recall more than 30\% of the categories. In comparison, the other ResNet-based models perform poorly that reach less than 20\% recall rate.
Furthermore, for RC@300, nearly 75\% of the categories in an image are captured by ViT-B/32, and RN50$\times$64 reaches about 60\% recall rate.
Given that the maximum detection number for LVIS v1.0 in OVOD is usually set to 300 (objects), ViT-B/32 can at most help detect 75\% of the objects if all the objects have different categories and 50\% if every 3 of the objects share the same category.
This provides substantial knowledge of categories to help the detector build the representation for multiple object detection.
Therefore, the CLIP image encoder is able to capture multiple categories in an image at the same time with relatively high accuracy and provide substantial knowledge of categories for the detector during image-level alignment.


\section{Conclusion}
In this work, 
we introduce a one-stage detector GridCLIP,
    which exploits the CLIP representation space to supplement the knowledge on undersampled categories in downstream detection datasets.
GridCLIP optimizes generalizable knowledge from the CLIP
  pretrained representation to more fine-grained localized mapping,
 by simultaneously
learning a localized grid-level CLIP
mapping to  base categories (Grid-level Alignment) and a holistic
image-level knowledge distillation to  base and novel
categories in a target
object detection domain (Image-level Alignment).
In our experiments, we verify that GridCLIP 
suffers less from long-tail distributions with the help of both grid-level and image-level alignments,
reaching comparable performance 
on the LVIS v1.0 benchmark \jy{with higher training and inference speed.}

{\small
\bibliographystyle{ieee_fullname}
\bibliography{GridCLIP}
}

\end{document}




\section{Implementation Details}
For data augmentation, a training image 
is first flipped with a probability of 0.5 
and then resized with the long edge set to
1333 and the short edge to 800 pixels, 
followed by the normalization with the mean and standard deviation in CLIP. 
For the input image of the fixed CLIP image encoder in image-level alignment, 
we follow the original processing in CLIP. 
When comparing with other SOTA approaches, we adopt multi-scale training with the sizes of
    (1333, 640), (1333, 672), (1333, 704), (1333, 736), (1333, 768), (1333, 800) similar to ViLD and random cropping augmentation where the cropped edge is not less than 0.5 of the original edge. 
    
For the training process,
     GridCLIP is trained for a 2$\times$ (24 LVIS epochs) schedule.
     GridCLIP uses AdamW optimizer with the learning rate as 1e-4
    and the decay weight as 1e-4, where the learning rate of the CLIP
    image encoder is 0.1 of the other parameters to better preserve the
    pre-trained weights.
The learning rate multiplied by 0.1 at epoch 16 and 22. We use linear warmup for the first 500 iterations, from learning rate 1e-3. 
    GridCLIP also adopts gradient clipping with 
    a max $L_2$ norm of 0.1 
    similar to that adopted in DenseCLIP. 
    Repeat factor sampling~\cite{gupta2019lvis} is adopted as in ~\cite{gu2021open,zhou2022detecting}. 
The batch size is 16.
During the inference stage, the maximum number of detection objects per image is 300,  and the threshold of the classification score is set to be 0.05 as default in MMDetection.
The IOU threshold of Non-Maximum Suppression is 0.5.

{\small
\bibliographystyle{ieee_fullname}
\bibliography{GridCLIP}
}